\documentclass{article}

\usepackage{arxiv}

\usepackage[utf8]{inputenc} 
\usepackage[T1]{fontenc}    
\usepackage{hyperref}       
\usepackage{url}            
\usepackage{booktabs}       
\usepackage{amsfonts}       
\usepackage{nicefrac}       
\usepackage{microtype}      
\usepackage{lipsum}		
\usepackage{graphicx}
\usepackage{natbib}
\usepackage{doi}
\usepackage{amsmath}
\usepackage{fancyhdr} 
\usepackage{multirow}

\title{Bifurcated Remaining Useful Life Prediction: A Hybrid Approach for Realistic Uncertainty Characterization}

\author{ 
    Xabier Belaunzaran \\
    Fundación Vicomtech \\
    Basque Research and Technology Alliance (BRTA)\\
    Mikeletegi 57, Donostia-San Sebastian, 20009, Spain \\
    \texttt{xbelauzaran@vicomtech.org} \\
    \And
    Antonio Nappa \\
    Fundación Vicomtech \\
    Basque Research and Technology Alliance (BRTA)\\
    Mikeletegi 57, Donostia-San Sebastian, 20009, Spain \\
    \texttt{anappa@vicomtech.org} \\
    \And
    Arkaitz Artetxe \\
    Fundación Vicomtech \\
    Basque Research and Technology Alliance (BRTA)\\
    Mikeletegi 57, Donostia-San Sebastian, 20009, Spain \\
    \texttt{aartetxe@vicomtech.org} \\
    \And
    Basilio Sierra \\
    University of the Basque Country (UPV/EHU) \\
    Donostia-San Sebastián, 20009, Spain \\
    \texttt{b.sierra@ehu.eus} \\
}

\begin{document}

\maketitle

\begin{abstract}
This study presents a novel hybrid prognostic framework for uncertainty-aware Remaining Useful Life (RUL) estimation in turbofan engines using the NASA C-MAPSS dataset. The framework employs a state-aware strategy that bifurcates the engine's operational lifespan into `healthy' and `degraded' regimes. An LSTM-based autoencoder, trained strictly on nominal data ($RUL > 150$ cycles), monitors reconstruction error to act as a robust state classifier. For the healthy regime, a Conditional Weibull Survival Analysis is used for Mean Residual Life estimation. For the degraded regime, a Probabilistic Neural Network with Monte Carlo Dropout captures both aleatoric and epistemic uncertainties. Rather than using rigid binary labels, a calibrated sigmoid function converts the autoencoder’s output into continuous state probabilities, dynamically weighting the final ensemble prediction. The primary strength of this framework is its generation of physically consistent uncertainty bands, yielding high-confidence predictions near end-of-life while accurately reflecting the inherent variance of early operation, providing a robust tool for risk-informed maintenance.
\end{abstract}


\section{Introduction}
\label{sec:introduction}
In the landscape of Prognostics and Health Management (PHM), estimating the Remaining Useful Life (RUL) of a system or component is crucial for transitioning from reactive to predictive maintenance strategies, thereby minimizing downtime and optimizing maintenance costs \cite{HUANG202478}. In safety-critical sectors such as heavy-manufacturing, automotive, and aviation, accurately predicting a component's RUL is vital; beyond safeguarding significant financial investments, it directly enhances operational safety and reliability of complex machinery \cite{LI2024111120}.

Historically, statistical population-level models and physics-based approaches, which require comprehensive domain knowledge of failure mechanisms, have been employed for RUL prediction \cite{Sutharssan}. In this context, the adoption of Artificial Intelligence (AI) and Deep Learning (DL) has revolutionized the PHM field. Unlike traditional approaches, asset-specific data-driven can automatically extract latent features from high-dimensional sensor data, enhancing predictive ability. Hybrid methods combining different approaches have also shown great performance \cite{Kimetal2017}.

Nevertheless, the high complexity of modern systems makes reliable prediction difficult. In real-world operations, where noise and varying operating conditions are ubiquitous, a point estimate without a confidence interval can be misleading and risky \cite{NEMANI2023110796}. Therefore, incorporating Uncertainty Quantification (UQ) into these predictions is becoming extremely important for decision-making, as it allows operators to understand and evaluate the risk associated with a prediction and make informed decisions. Although significant work has been dedicated to enhance the accuracy of RUL estimations, in uncertainty quantification research there is still room for improvement. While simple UQ methods in literature start to apply to the PHM field, there remains a need for robust methods to quantify the uncertainty associated with ML and complex hybrid methods \cite{SALINASCAMUS2025113015}.

In this paper, we address this gap by proposing a robust framework for RUL prediction with integrated UQ. The approach is validated using the widely recognized NASA C-MAPSS 2008 turbofan engine degradation dataset, which represents one of the main benchmark datasets on the PHM field. Specifically, we introduce a novel hybrid approach that overcomes the limitations of the rigid Piecewise Linear (PwL) target assumptions commonly used in literature. By combining an autoencoder with a Probabilistic Long-Short-Term-Memory (LSTM) and a Conditional Weibull survival analysis, our framework captures both aleatoric and epistemic uncertainties. 

The main contributions of this paper are summarized as follows:
\begin{itemize}
    
    \item We introduce a custom Asymmetric Negative Log-Likelihood loss function to train the Probabilistic LSTM network. This model explicitly quantifies predictive uncertainty while penalizing dangerous late predictions, aligning with the safety-critical reality of aviation maintenance.
    \item As an alternative to the PwL function we propose a multi-step approach: (i) an autoencoder-based classification into healthy and degraded regions. (ii) RUL estimation using Weibull Survival Analysis in the healthy region and Probabilistic LSTM network in the degraded one. (iii) Fusion of predictions and uncertainty sources of both models through probability weights guided by the autoencoder’s labeling.

\end{itemize}

The remainder of this paper is organized as follows: Section 2 reviews the state of the art in data-driven RUL prediction and Uncertainty Quantification. Section 3 describes the C-MAPSS benchmark dataset and its specific challenge score function. Section 4 presents the theoretical background of the key ideas. Section 5 details the proposed hybrid methodology, including the autoencoder segmentation, the Conditional Weibull model, and the Probabilistic LSTM. Section 6 presents the experimental results and evaluates the uncertainty metrics. Finally, Section 7 concludes the paper and discusses future research directions.

\section{State of the Art}
\label{sec:state_of_the_art}

In the field of prognostics, traditional approaches have relied on statistical survival analysis \cite{survival} to estimate system longevity. Methods such as Cox Proportional Hazards models \cite{cox} and Weibull distribution analysis \cite{weibull_original} have been widely used to model the probability of failure over time. These approaches operate at the population level, leveraging historical failure data across multiple assets to estimate general degradation trends and associated uncertainty.

More recently, data-driven approaches have come to dominate Remaining Useful Life (RUL) prediction, particularly on the C-MAPSS dataset. These methods leverage architectures capable of capturing temporal dependencies in high-dimensional sensor data. Due to the sequential nature of the input sensor data and the architecture's ability to capture long-term temporal dependencies, the LSTM is one of the most successful models used on the C-MAPSS dataset \cite{zheng2017long}. Other architectures include Convolutional Neural Networks (CNN) \cite{li2018remaining}, attention-based models \cite{LIU2022108330}, and hybrid models \cite{Elsherif2025}. In contrast to statistical approaches, these architectures are inherently asset-specific, learning degradation patterns directly from individual run-to-failure trajectories.

A fundamental challenge in training supervised RUL models is that observable degradation is minimal during early operational cycles. This introduces noise, as models attempt to map near-static inputs to strictly decreasing RUL targets. To address this issue, Heimes introduced the Piecewise Linear (PwL) target function \cite{heimes2008recurrent}, which caps the maximum RUL during the initial stable phase at a constant value ($irul$), typically between 125 and 130 cycles \cite{LIU2022108330}. This approach ensures that the model focuses on learning degradation patterns only when they become physically meaningful. Indeed, most state-of-the-art methods adopt this strategy \cite{MITICI2023109199, asif2022deep}.

Specific hybrid architectures have already been proposed to address this challenge. For instance, Arunan et al. trained a model using an engine-specific piecewise function, where the upper limit is determined through change-point detection via canonical variate analysis \cite{arunan2024change}. In this framework, online predictions begin just once degradation is detected. Other hybrid approaches include the work by Cui et al., who combine a Cox proportional hazards model with an XGBoost predictor \cite{CUI2026111451}, and Song et al. (2018), who utilize an autoencoder for feature extraction followed by a bidirectional LSTM for prediction \cite{Song2018}.

In parallel, there is a growing shift from deterministic point estimates toward approaches that explicitly quantify uncertainty, which is critical for risk-aware decision-making in aviation maintenance. Reliable uncertainty quantification (UQ) requires capturing both aleatoric and epistemic uncertainty \cite{Gal2016UncertaintyID}. 

Statistical survival models are well-suited for capturing population-level uncertainty, as their predictions are derived from the variability observed across a fleet of assets. Recent studies have introduced analytical methods to further quantify these dynamics. For instance, Younus and Cai implemented Weibull Reliability Modeling with integrated Uncertainty Quantification (UQ) \cite{Weibull_UQ}, while Dersin and Rocchetta utilized time transformations to measure predictive uncertainty \cite{DERSIN2026111730}.

Regarding data-driven approaches based on asset-specific temporal patterns, a variety of methods have been proposed for this purpose \cite{NEMANI2023110796}, including variational inference in Bayesian Neural Networks (BNNs) \cite{peng2020bayesian}, Deep Gaussian Processes \cite{biggio}, Monte Carlo (MC) Dropout \cite{mc_dropout, ochella}, and ensemble-based approaches. Comparative studies on the C-MAPSS dataset further evaluate these techniques \cite{BASORA2025110513, XUAN2023116}, including quantile regression and hyper-deep ensembles. However, state-of-the-art methods that incorporate both PwL targets and uncertainty quantification tend to exhibit low predictive uncertainty during early operational stages, where degradation signals are not yet observable. Since the model learns to output a constant value determined by the PwL cap, this behavior reflects low epistemic uncertainty, but it does not accurately represent the true uncertainty in the system’s health state.

To address this limitation, we propose a hybrid framework that decouples the modeling of healthy and degrading regimes. Instead of relying on a single model or a rigid PwL constraint, our approach employs an autoencoder to gate predictions between two complementary components: a statistical survival model for healthy states and a probabilistic LSTM for degradation phases. This design leverages the strong predictive performance of LSTM-based models while improving uncertainty estimation in early stages by incorporating survival-based predictions. The method combines population-level statistical uncertainty calculations with asset-specific methods, resulting in more realistic uncertainty bounds, that are wider in the earlier stages and decrease along the trajectories.

\section{CMAPSS 2008 Challenge}

This study utilizes the widely recognized NASA C-MAPSS turbofan engine degradation dataset \cite{saxena2008damage}, which serves as a standard benchmark in the PHM field. The dataset was generated on 2008 using the Commercial Modular Aero-Propulsion System Simulation (C-MAPSS) tool \cite{frederick2007user}.  It consists of multivariate time-series data featuring 21 sensor measurements and 3 operational settings. The data covers complete run-to-failure training trajectories and truncated testing trajectories across four sub-datasets with varying levels of complexity. Specifically, FD001 operates under a single condition with a unique failure mode; FD002 and FD004 include six different operational conditions; and FD003 and FD004 involve two distinct failure modes.

To evaluate model performance, we utilize the standard Root Mean Square Error (RMSE) and Mean Absolute Error (MAE) to measure overall point-estimate accuracy, alongside the official C-MAPSS asymmetric scoring function, defined as:

\begin{equation}
    S = \sum_{i=1}^{N}s_i \quad  \quad with \quad 
    s_i = \begin{cases} 
e^{-\frac{d_i}{13}} - 1, & \text{for } d_i < 0 \\ 
e^{\frac{d_i}{10}} - 1, & \text{for } d_i \ge 0 
\end{cases}
\label{eq:score}
\end{equation}

where $ d_i = \hat{RUL}_i - RUL_i $ represents the prediction error. This asymmetric function penalizes late predictions ($d_i \geq 0$ ) significantly more heavily than early predictions ($d_i \leq 0$). This explicitly reflects the safety-critical reality of aviation maintenance, where overestimating remaining life can lead to catastrophic failure, whereas underestimating merely results in premature maintenance.

\section{Background}
\label{sec:background}

\subsection{Autoencoder residual modeling}
\label{sec:autoencoder_theory}

The use of autoencoder-based signal reconstruction is a common strategy in prognosis for constructing health-state indicators. The core premise relies on defining a baseline for normal engine behavior and subsequently measuring how far incoming sensor sequences deviate from this healthy baseline \cite{gonzalez}. By training the autoencoder exclusively on healthy sensor sequences, the model learns to compress and reconstruct the underlying patterns. When the model encounters anomalous or degraded data, the reconstruction quality diminishes. This increase in the residual error serves as a direct proxy for mechanical degradation.

To quantify the transition from healthy to degraded states, a statistical boundary based on the reconstruction performance is established over the healthy training set. Assuming the reconstruction MAE errors follow a normal distribution, the mean ($\mu$) and standard deviation ($\sigma$) of these errors are calculated across the healthy training set. A critical degradation threshold is defined in this distribution to represent the extreme upper limit of normal behavior.

Then, the degradation state of the system can be evaluated by reconstructing the signal with this autoencoder, and passing the reconstruction MAE values through a calibrated Sigmoid function centered exactly at the threshold $T$:

\begin{equation}
    P(degraded) = \frac{1}{1+e^{-k(MAE-T)}}
    \label{Eq:Probability}
\end{equation}

Here, $k$ is a scaling parameter that controls the steepness of the transition. The resulting output smoothly transitions from near 0 to near 1 as the engine degrades. When the MAE of the reconstructed window exactly equals the threshold $T$, the function outputs a degradation probability of $0.5$. A perfect reconstruction ($MAE = 0$) does not yield an absolute 0, but rather a value asymptotically close to it, reflecting a natural baseline uncertainty. It is important to clarify that the quantity $P(degraded)$ serves as a heuristic gating weight rather than a formal probability.

\subsection{Weibull Survival Analysis}

The Weibull distribution is a continuous probability function, widely used to model time to failure events for survival analysis. It is defined by its scale parameter $\lambda$ and shape parameter $k$. Because turbofan engines exhibit an increasing failure rate over time due to wear and degradation, the distribution fits with a shape parameter $k>1$. The baseline survival function is given by $S(t)=e^{-(t/\lambda)^k}$ \cite{weibull_original}.

In real-world prognostics, predictions occur at a current operating cycle $t_0$, meaning the engine has already survived to $t_0$ without failing. Thus, predicting its RUL is a problem of conditional probability. The probability of surviving an additional duration x (the RUL), given survival up to $t_0$, is:

\begin{equation}
    S(x|t_0) = \frac{S(t_0 + x)}{S(t_0)} = e^{\left(\frac{t_0}{\lambda}\right)^k - \left(\frac{t_0 + x}{\lambda}\right)^k}
    \label{Eq:CondSurvival}
\end{equation}

Using this conditional survival function, the expected RUL at time $t_0$, formally known as the Mean Residual Life (MRL), is calculated by integrating the conditional survival probabilities over all possible future times $x$:

\begin{equation}
    \mu_{Weib}(t_0) = \int_{0}^{\infty} S(x|t_0) dx = \frac{\int_{0}^{\infty} S(t_0 + x) dx}{S(t_0)}
    \label{Eq:MRL}
\end{equation}

The uncertainty of the MRL can be related to the variance of the conditional distribution, which is calculated by numerically evaluating the second moment of the conditional distribution, $E[X^2|t_0] = 2 \int_{0}^{\infty} x \cdot S(x|t_0) dx$. The conditional variance $\sigma^2_{Weib}$ is therefore obtained through the standard variance identity:

\begin{equation}
    \sigma_{Weib}^2(t_0) = E[X^2|t_0] - \mu^2_{Weib}(t_0)
    \label{Eq:WeibullVariance}
\end{equation}

\subsection{Probabilistic framework for Uncertainty Calculation}
\label{sec:Bayesian_UQ}
\subsubsection{NLL Loss function for Aleatoric Uncertainty}

In Probabilistic Neural Networks, the Negative Log-Likelihood (NLL) replaces standard deterministic loss functions like Mean Squared Error (MSE). The network's objective shifts from predicting a single value to predicting the parameters ($\mu$ and $\sigma^2$) of a Gaussian distribution that best describes the target data. The standard NLL maximizes the likelihood that the observed ground-truth RUL belongs to the predicted distribution. Representing this ground-truth RUL as $y$, the NLL loss is then defined as:

\begin{equation}
\label{eq:NLL}
    NLL = \frac{1}{2} \log \sigma^2 + \frac{(y-\mu)^2}{2\sigma^2}
\end{equation}

By minimizing the NLL, the loss function explicitly penalizes the model for being confidently wrong. If the model's mean prediction is far from the true target, it must increase its predicted variance to avoid a massive loss penalty. This forces the model to self-report the aleatoric (data-driven) uncertainty inherent in the degradation process.

\subsubsection{Monte Carlo Dropout for Epistemic Uncertainty}
\label{sec:MC}
While the NLL loss function captures aleatoric uncertainty, it does not account for epistemic (model) uncertainty. To evaluate this, MC Dropout is utilized. Theoretical research \cite{mc_dropout} demonstrated that leaving dropout active during the inference phase is mathematically equivalent to performing approximate variational inference on a probabilistic model.

By passing the exact same input sequence through the model $M$ times, turning off a random subset of neurons during each forward pass, an ensemble of different sub-networks is sampled. The variance across the predicted means ($\mu$) of these $M$ forward passes directly quantifies the epistemic uncertainty. The overall predictive uncertainty ($\sigma_{total}^2$) is calculated as the sum of the mean aleatoric uncertainty and the epistemic uncertainty:

\begin{equation}
    \label{Eq:MC}
    \sigma_{total}^2 = \underbrace{\frac{1}{M}\sum_{m=1}^{M}\sigma_m^2}_{\text{Aleatoric}} + \underbrace{\frac{1}{M}\sum_{m=1}^{M}\mu_m^2 - \left(\frac{1}{M}\sum_{m=1}^{M}\mu_m\right)^2}_{\text{Epistemic}}
\end{equation}

\subsubsection{Uncertainty Evaluation Metrics}

The primary aim of the proposed research is to reliably quantify the uncertainty of RUL predictions. Therefore, in addition to traditional deterministic performance metrics like RMSE and the C-MAPSS challenge score, specific probabilistic metrics are required to evaluate the quality of the generated prediction intervals \cite{ZHAN2024110383}. For a given confidence level (e.g., 95\%), the upper ($U_i$) and lower ($L_i$) bounds are calculated using the total predicted variance.

To assess the quality of these uncertainty bands, we utilize the Proportional Index ($N/P$). This metric balances two competing factors: the Prediction Interval Coverage Probability (PICP), which measures the proportion of true target values falling within the predicted bands, and the Normalized Mean Prediction Interval Width (NMPIW), which measures the average width of these intervals. While a wider interval artificially increases coverage (PICP), it reduces practical usefulness (increasing NMPIW). The Proportional Index provides a comprehensive evaluation by capturing this trade-off, where a highly effective probabilistic model minimizes this ratio:

\begin{equation}
    N/P = \frac{NMPIW}{PICP}
\end{equation}

\section{Methodology}
\label{sec:methodology}
\subsection{Proposed Framework Overview}

A primary challenge when predicting RUL on the C-MAPSS dataset is the lack of observable degradation in sensor readings during the initial operating cycles. In this early phase, sensor data remains largely static, yet the ground-truth RUL strictly decreases with each passing cycle. To prevent neural networks from becoming confused by mapping identical sensory inputs to a wide variety of target RUL values, the literature widely adopts a Piecewise Linear (PwL) target function, capping the maximum initial RUL at a constant value ($irul$) (Section \ref{sec:state_of_the_art}).

While utilizing this PwL function undeniably improves model convergence and boosts overall performance metrics, it acts primarily as a mathematical constraint. The predictive model learns to output the predefined $irul$ constant when fed healthy data, but it does not genuinely improve the model's capacity to learn underlying physical degradation patterns during this phase. Fundamentally, deep learning models struggle to extract dynamic degradation signals where none yet exist.

To address this inherent characteristic of the data, we propose a novel, two-stage hybrid framework. Rather than forcing a single neural network to learn both the static healthy phase and the dynamic degraded phase simultaneously, our methodology separates these regimes. The proposed approach consists of three main components:

\begin{enumerate}
    \item \textbf{Degradation-state gating via autoencoder:} An unsupervised LSTM-based autoencoder is trained to classify input windows, transitioning from a binary `healthy' or ´degraded' classification to a continuous probability of degradation. 
    \item \textbf{Conditional Weibull survival analysis:} For the healthy phase, where predictive models lack sufficiently informative degradation signal, the RUL is estimated using Conditional Weibull survival analysis. This leverages the known survival time of the engine to provide a physically meaningful expected life.
    \item \textbf{Probabilistic LSTM:} For the degraded phase, a Probabilistic LSTM is trained exclusively on deteriorating data. This eliminates the noise of the stable phase and allows the model to predict RUL while quantifying both aleatoric and epistemic uncertainty.
\end{enumerate}

\begin{figure}[t]
  \centering
  \includegraphics[scale=.3]{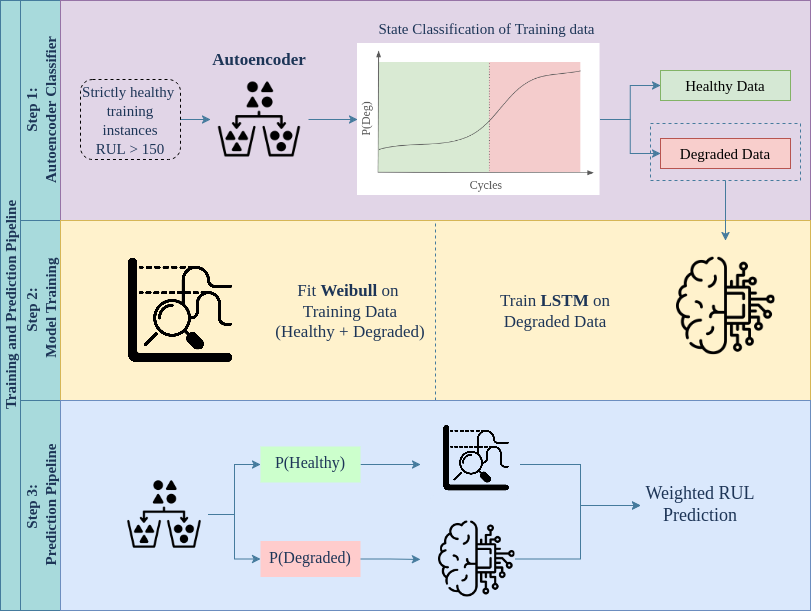}
  \caption{Illustration summarizing the proposed pipeline. The autoencoder is trained with strictly healthy instances, and classifies the instances on the training set as healthy or degraded. Then the Weibull distribution is fitted with training data and the LSTM with exclusively degraded training data. The final prediction is performed as a weighted ensemble of both approaches, where the prediction weights are provided by the output degradation probabilities of the autoencoder.}
  \label{fig:pipeline}
\end{figure}

The final RUL prediction for any given window is formulated as a weighted average of the statistical (Weibull) and predictive (Probabilistic LSTM) strategies, where the weights correspond directly to the degradation probabilities output by the autoencoder. The described pipeline is summarized in Figure \ref{fig:pipeline}.

\subsection{Data Preprocessing Pipeline}

Prior to model training, the dataset undergoes a systematic preprocessing pipeline to mitigate noise, reduce dimensionality, and format the temporal data for deep learning:

\begin{itemize}
    \item \textbf{Feature Selection}: Following standard practices established in the literature \cite{zheng2017long}, sensors that record constant values throughout the trajectories are discarded, reducing the input dimensionality.
    \item \textbf{Signal Smoothing}: A moving median filter with a sliding window of 12 cycles is applied to each sensor. This robustly eliminates high-frequency noise and sudden spikes while preserving critical monotonic degradation trends.
    \item \textbf{Normalization}: The filtered data is scaled to the [0, 1] range using Min-Max normalization, ensuring stable LSTM convergence. To prevent data leakage, scaling parameters are computed exclusively from the training set and applied to the validation and testing sets.
    \item \textbf{Window Splitting and Dataset Partitioning}: To leverage the temporal learning capabilities of the deep learning models, the continuous time-series data is transformed into 2D matrices using a sliding window approach \cite{li2018remaining}. For each dataset, the window size has been set to the value matching the minimum trajectory length in the testing set. Finally, the original training trajectories are partitioned into a new training set (80\%) and a validation set (20\%). This split is strictly enforced at the engine level, ensuring that all windows belonging to a specific engine are kept exclusively within either the training or validation set.
\end{itemize}

In datasets FD002 and FD004, which involve multiple operating conditions, a clustering algorithm\footnote{Specifically \textit{K-Means} from \textit{scikit-learn}} was employed to classify the six operating regimes. Subsequently, cluster-specific normalization was applied to each group. Since the operating conditions vary over time, this procedure necessitates performing the signal smoothing step after normalization to ensure that transition noise between regimes is properly handled.

\subsection{Health-State Classification via LSTM Autoencoder}

The first stage of the proposed framework involves training an anomaly detection model capable of distinguishing between healthy' and `degraded' operational states. This is achieved through the residual modeling framework described in Section \ref{sec:autoencoder_theory}.

First, to establish a robust baseline of healthy behavior, early-life operational data from the C-MAPSS engines was isolated. The training subset was filtered to include only instances where the ground-truth RUL exceeded 150 cycles. While this 150-cycle threshold is a heuristic, it is intentionally set higher than the widely accepted 125-cycle piecewise RUL standard. This ensures that the majority of the selected input windows represent genuinely healthy, steady-state operations.

Then, input windows generated from this healthy subset are used to train the unsupervised Sequence-to-Sequence LSTM autoencoder. The input sliding window is first processed by an Encoder LSTM, which extracts temporal features and compresses them into a single, static latent vector at the bottleneck. A RepeatVector layer then duplicates this compressed representation across all time steps, feeding it into a Decoder LSTM. Finally, a TimeDistributed Dense layer projects the decoder's output back into the original multivariate sensor space, reconstructing the time-series step-by-step.

The anomaly threshold is defined as $T = \mu + 3\sigma$ , where $\mu$ and $\sigma$ are derived from the reconstruction MAE of the healthy training set. Selecting a high $3\sigma$ boundary minimizes false alarms caused by standard signal noise while remaining highly sensitive to equipment wear. For the Sigmoid mapping in Eq.\ref{Eq:Probability}, the scaling parameter is set to $k = 100$.  The parameter values have been heuristically selected.

\subsection{Conditional Weibull Survival Analysis for Healthy States}

To generate realistic predictions during the early operational phase prior to observable degradation, a Conditional Weibull Survival Analysis is employed. Unlike the static Piecewise Linear \textit{irul} target assumption, this approach leverages the known survival time of the engine to provide a statistically rigorous expectation of remaining life. 

To integrate this statistical approach into our UQ framework, both the mean $\mu_{Weib}$ and the variance $\sigma_{Weib}^2$ of the conditional distribution are calculated, using Equations \ref{Eq:MRL} and \ref{Eq:WeibullVariance}. Thereby the statistical model provides a complete probabilistic prediction for the healthy phase, and it mirrors the output format by the Probabilistic LSTM during the degraded phase, allowing both models to be fused using the autoencoder's probability weights.

\subsection{Predictive Model for Degraded States}

The second predictive strategy employs a deep learning model trained exclusively on the degraded windows identified by the autoencoder. By gating the data, this approach effectively eliminates the noise and target confusion introduced by the static healthy phase, allowing the neural network to focus entirely on modeling active deterioration.

For this phase, a Probabilistic LSTM network is implemented. As discussed in Section \ref{sec:state_of_the_art}, a prognostic framework should capture both sources of predictive variance. By doing so, the Probabilistic LSTM enables engineers to comprehensively assess the model's confidence, supporting risk-averse and safety-critical maintenance decisions. 

The initial layer takes the two-dimensional temporal window of the input sensors ($Window Size \times Features$) as its input, and is followed by a Convolutional Layer that captures spatial dependencies. The core of the model architecture is then constructed by interleaving LSTM layers with Dropout layers.  The number of these LSTM and Dropout layer pairs is treated as a hyperparameter. Concurrently, the absolute operational cycle number of the engine at the time of prediction, which corresponds to the last cycle count of the input sliding window, is introduced as an exogenous scalar variable. Because the sliding window only provides a localized 30-cycle snapshot of degradation, concatenating this absolute cycle number with the flattened output of the final LSTM layer provides the network with additional temporal context that grounds the localized sensor patterns within the broader lifespan of the engine.
The initial layer takes the two-dimensional temporal window of the input sensors ($Window Size \times Features$) as its input, and is followed by a Convolutional Layer that captures spatial dependencies. The core of the model architecture is then constructed by interleaving LSTM layers with Dropout layers.  The number of these LSTM and Dropout layer pairs is treated as a hyperparameter. Concurrently, the absolute operational cycle number of the engine at the time of prediction, which corresponds to the last cycle count of the input sliding window, is introduced as an exogenous scalar variable. Because the sliding window only provides a localized 30-cycle snapshot of degradation, concatenating this absolute cycle number with the flattened output of the final LSTM layer provides the network with additional temporal context that grounds the localized sensor patterns within the broader lifespan of the engine.

This fused representation passes through an intermediate Dense layer, followed by another Dropout layer. The final output is a Dense layer with two distinct neurons, representing the predicted mean ($\mu$) and variance ($\sigma^2$) of the RUL distribution.

The number of hidden units, $N$, is consistent across all LSTM layers, while the intermediate Dense layer contains $N/2$ neurons. These architectural parameters, along with the learning rate and network depth, were tuned during hyperparameter optimization over 30 trials using a Bayesian Sampler\footnote{KerasTuner with Bayesian Optimization has been employed.}, each with 100 epochs and early-stopping option. Table~\ref{tab:hyperparameters} summarizes the explored parameter ranges and the optimal values identified during this search.

\begin{table}[t] \small 
    \begin{center} 
    \caption{Hyperparameter values optimized during network training. The range indicates the explored values, and the step specifies the discrete increments between candidates.}
        \label{tab:hyperparameters}
        \begin{tabular}{c||c|c|c}
        Hyperparameter & Range & Step & Best value \\ 
        \hline
        Nº layers & $\{1,2,3\}$ & $1$ & $2$ \\  
        Nº neurons layer & $[32, 128]$ & $32$ & $128$ \\
        Dropout Rate & $[0.05, 0.4]$ & $0.05$ & $0.15$ \\
        Learning Rate & $[10^{-4}, 3\cdot10^{-3}]$ & Log & $2.16\cdot10^{-3}$ \\
        \end{tabular}
	\end{center}
\end{table}

\subsubsection{Asymmetric Negative Log-Likelihood Loss Function}

In the C-MAPSS challenge, predictions are evaluated using an asymmetric score function that heavily penalizes late predictions (Eq. \ref{eq:score}). However, this standard exponential score function cannot be used directly as a loss function for gradient-based training because its exponential growth causes exploding gradients during backpropagation. 

To train a model that favors early predictions while simultaneously estimating predictive uncertainty, we propose a custom Asymmetric Negative Log-Likelihood ($AsNLL$) function. This function applies a weighting factor $C$ to the squared error component of the NLL (Eq. \ref{eq:NLL}) for dangerously late predictions ($\mu > y$), preserving the mathematically necessary precision and variance terms required for proper NLL optimization:

\begin{equation}
    AsNLL = \frac{1}{2} \log \sigma^2 + \frac{C \cdot (y-\mu)^2}{2\sigma^2}  
\end{equation}
\begin{equation*}
    \text{where} \quad C = \begin{cases} 3, & \text{for } \mu > y \text{ (Late Prediction)} \\ 1,  & \text{for } \mu \le y \text{ (Early Prediction)} \end{cases}
\end{equation*}

While an exhaustive optimization of the penalty factor $C$ remains as a future challenge, the value has been optimized graphically analyzing the effect of this factor, training LSTM models with different $C$ values. The value of $3$ has been selected as the trade-off value that effectively mitigates the risk of overestimation during the critical mid-range of operation, without introducing the excessive conservative bias in the healthy states that characterizes higher penalty factors.

Following Probabilistic Framework for UQ described in Section \ref{sec:Bayesian_UQ}, while this AsNLL function serves to capture aleatory uncertainty on data, MC approach described in Section \ref{sec:MC} is employed to calculate epistemic uncertainty. The total overall predicting uncertainty is obtained by summing both components (Eq \ref{Eq:MC}).

\subsection{Weighted sum}

The final stage of the proposed framework combines the predictions from the healthy-state statistical model and the degraded-state deep learning model into a single output. One option to sum the distributions from both models would be to use Bayes' Theorem. This strategy would automatically give more weight to the model with less uncertainty, which, in the proposed framework, would result in neglecting the degradation information provided by the autoencoder. Indeed, the weights would be leveraged by the calculated uncertainties, but, because the predictive model is trained to work specifically in the degraded range, it is not guaranteed that the model's predictions and uncertainty quantification will be reliable within all the range.  We therefore propose a fusion governed by this continuous degradation probability $P_d$ proportioned by the autoencoder.

To formulate the ensemble prediction, we assign dynamic weights to each model based on the operational state probability. The weight for the Probabilistic LSTM prediction is defined as $w_{d} = P_{d}$, while the weight for the Conditional Weibull prediction is its complement, $w_{h} = 1 - P_{d}$.

Because both models output probability distributions defined by a mean and a variance, the combined output represents a mixture distribution. The final ensemble mean RUL, $\mu$, is calculated as the expected value of this mixture, which simplifies to a standard linear combination of the individual model means:

\begin{equation}
    \mu = w_h \cdot \mu_{Weib} + w_d \cdot \mu_{LSTM}
\end{equation}

Calculating the total predictive uncertainty requires more than a simple weighted average of the variances. To accurately reflect the uncertainty, we must account for both the internal variance of each model and the dispersion between the models' conflicting predictions. Applying the Law of Total Variance for a mixture distribution, the final ensemble variance, $\sigma^2$, is computed as:

\begin{equation}
\begin{split}
    \sigma^2 &= w_h\sigma^2_{Weib} + w_d\sigma^2_{LSTM} \\
    &\quad + (w_h(\mu_{Weib}-\mu)^2+w_d(\mu_{LSTM}-\mu)^2)
\end{split}
\end{equation}

The first two terms represents the weighted internal uncertainty (aleatoric and epistemic) of the individual models. The second clustered term captures the divergence between the models' mean estimates. Consequently, when the autoencoder is highly uncertain about the health state and the two models predict vastly different RULs, the total uncertainty band widens. This naturally enforces a conservative, safety-first uncertainty quantification right at the critical degradation transition point.

\section{Results and Discussion}
\label{sec:results}
\subsection{Health-State Classification via Autoencoder}

This section evaluates the efficacy of the unsupervised autoencoder in establishing a dynamic health-state baseline. As detailed in the methodology, the autoencoder was trained exclusively on heuristically defined `healthy' instances (trajectories with an $RUL > 150$ cycles). Because the model is unsupervised, this initial validation qualitatively assesses whether the autoencoder's learned reconstructions physically align with expected engine degradation patterns. A quantitative validation of the overall predictive framework is presented in subsequent sections.

During the first phase, the autoencoder was trained on the healthy subset, and the MAE was calculated for the reconstruction of these same healthy training instances. 

\begin{figure}[t]
  \centering
  \includegraphics[scale=.33]{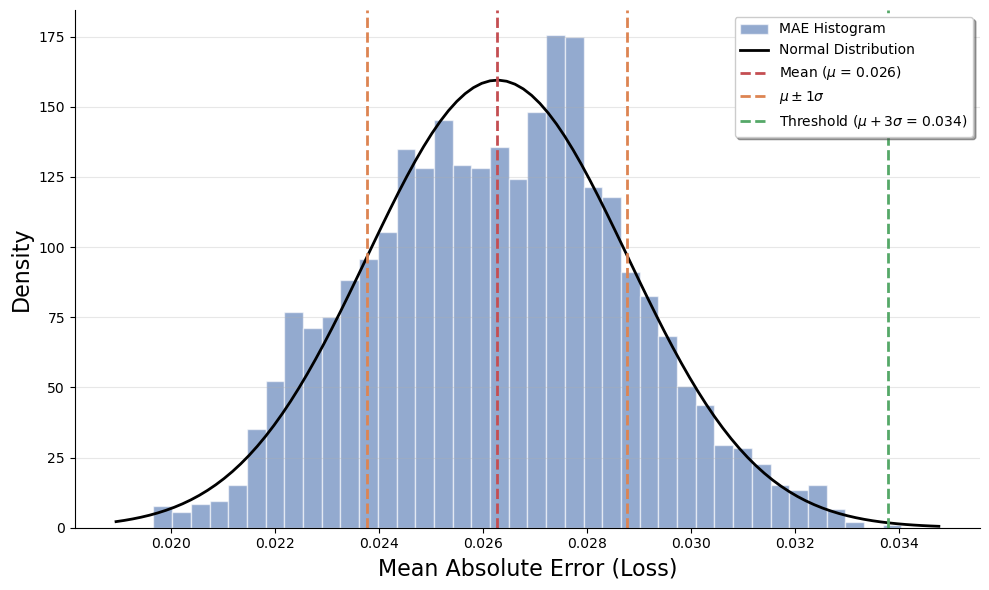}
  \caption{Distribution of MAE Loss data of the windows reconstructed with the autoencoder over the original windows of FD001 training data. The distribution has been calculated with the same healthy instances used to train the autoencoder.}
  \label{fig:autoencoder_training}
\end{figure}

Figure \ref{fig:autoencoder_training} illustrates the distribution of these MAE values in FD001 dataset. The reconstruction errors closely approximate a normal distribution, validating the statistical assumption underlying our threshold logic. The vertical dashed lines represent the mean ($\mu$), standard deviation ($\sigma$), and the critical $3\sigma$ boundary. Instances surpassing the threshold $T = \mu + 3\sigma$ (indicated by the green line) are flagged as non-healthy states. Notably, a small fraction of windows within this initial healthy subset ($RUL > 150$) exceeds the threshold $T$. Rather than being mere statistical anomalies, this reflects the physical reality that operating conditions vary, and some engines experience early-onset degradation.

\begin{figure}[t]
  \centering
  \includegraphics[scale=.33]{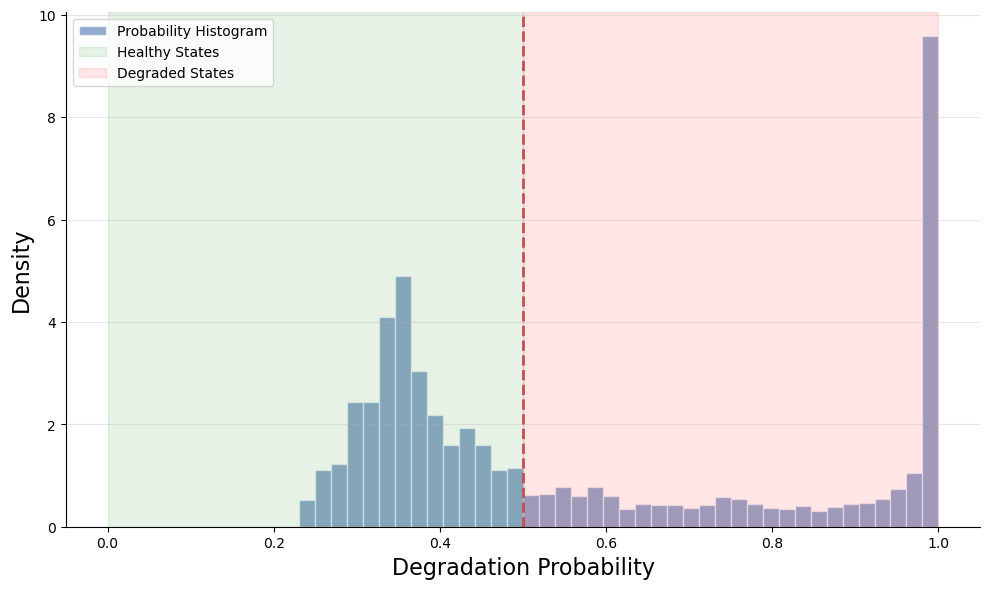}
  \caption{Distribution of the calibrated degradation probability for the complete training and validation sets of the FD001 dataset.}
  \label{fig:autoencoder_trainval_prob}
\end{figure}

Once trained and calibrated, the autoencoder was deployed across the complete training and validation sets (which include both fully healthy and degraded instances). Figure \ref{fig:autoencoder_trainval_prob} displays the degradation probabilities after passing the reconstruction MAE in this dataset through the calibrated Sigmoid function (Eq. \ref{Eq:Probability}). 

In stark contrast to the initial healthy distribution, Figure \ref{fig:autoencoder_trainval_prob} exhibits a significant proportion of instances (6,390 out of 17,731 windows, or roughly 36\%) classified as degraded. Furthermore, the high density of instances approaching a probability of 1.0  indicates the model's high confidence in detecting severe deterioration. Conversely, the broader spread at lower probabilities reflects the gradual nature of the transition from a healthy to a degraded state. The proportion of data points classified as degraded is comparable across the remaining subsets (36\%, 35\% and 26\% for datasets FD002, FD003 and FD004 respectively). However, a distinct shift in behavior is observed in datasets with multiple operating conditions, where the autoencoder demonstrates higher confidence in distinguishing between healthy and degraded states, resulting in fewer instances within the transition phase.

\subsection{Conditional Weibull Survival Analysis for Healthy States}

To establish the statistical baseline for the early-life operational phase, a Weibull distribution was fit\footnote{The distribution is fit using the \textit{SciPy} library in Python} to the run-to-failure trajectories of the training set. After defining the base survival function with the fitted scale and shape parameters, the conditional mean and variance for the testing samples were calculated utilizing Eqs. \ref{Eq:MRL} and \ref{Eq:WeibullVariance}, respectively. The number of elapsed cycles since the start of the trajectory ($t_0$) served as the conditioning input variable. 

\begin{figure}[t]
  \centering
  \includegraphics[scale=.3]{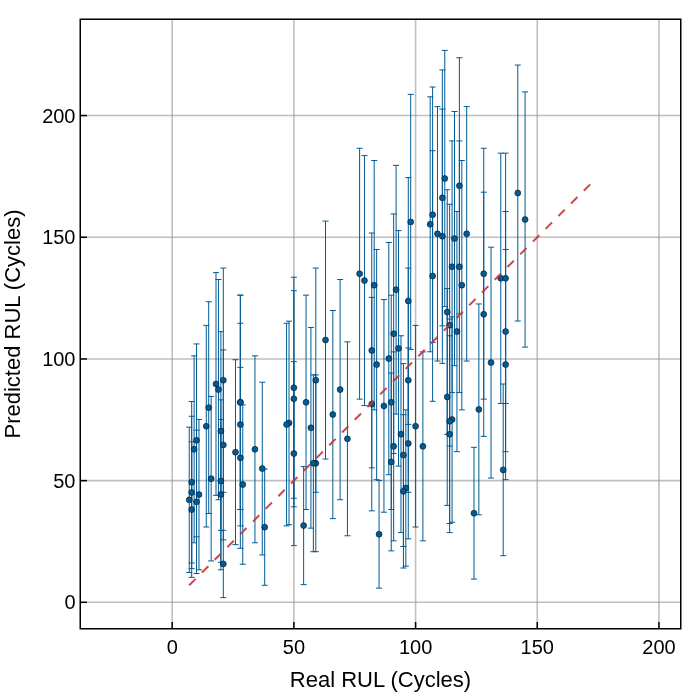}
  \caption{RUL prediction performed using Weibull Survival Analysis. Error bars represent the calculated standard deviation $\sigma$.}
  \label{fig:weibull_predictions}
\end{figure}

Figure \ref{fig:weibull_predictions} illustrates the results of these statistical predictions in FD001, plotting the predicted Mean Residual Life against the ground-truth RUL. The uncertainty bands in the plot directly represent the calculated conditional variance ($\sigma_{Weib}^2$).

As observed in the plot, the survival analysis yields wide confidence intervals for high RUL values, which accurately reflects the natural variance in the total lifespan of different engines. The slight positive correlation trend in the plot demonstrates that the conditional approach dynamically adjusts its predictions downwards as the engine accumulates cycles, providing a mathematically sound, decreasing expectation of life rather than a rigid constant. However, this approach in isolation does not offer optimal performance across all operational points. 

\subsection{Probabilistic LSTM Predictions for Degraded States}

\begin{figure}[t]
  \centering
  \includegraphics[scale=.3]{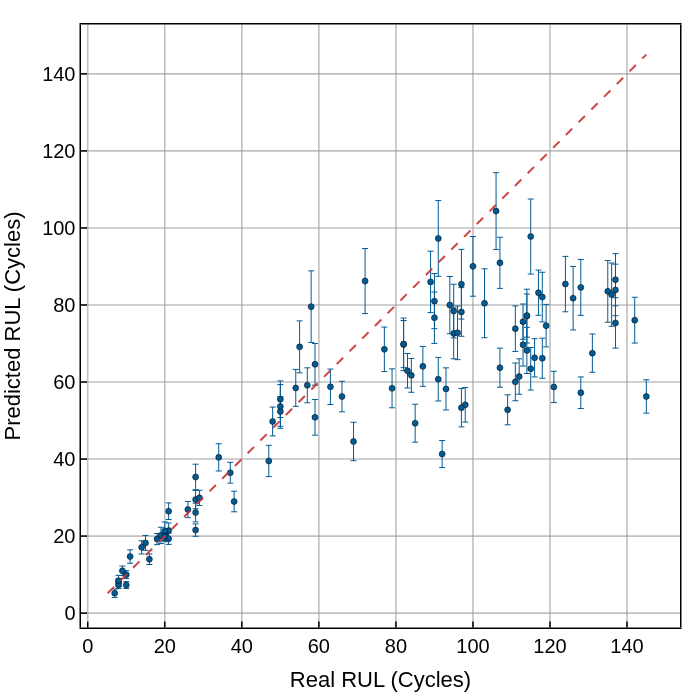}
  \caption{RUL prediction performed with the Probabilistic LSTM trained with only degraded instances. Error bars represent the calculated standard deviation $\sigma$.}
  \label{fig:lstm_degraded}
\end{figure}

To evaluate the deep learning component independently, Figure \ref{fig:lstm_degraded} displays the testing predictions of FD001 dataset generated by the Probabilistic LSTM, which was trained exclusively on the degraded instances identified by the autoencoder.

While model shows high precision for low RUL values, the most prominent feature of this plot is the distinct horizontal clustering of predictions in the zone of high true RUL values (the healthy phase). This behaviour, common to all the datasets, visualizes the fundamental limitation addressed in our methodology: when fed sensor data from a healthy engine, the deep learning model cannot distinguish between different true RUL values because the degradation signal has not yet manifested. Consequently, the network simply learns to output a constant value that minimizes its loss for that region. This explicitly validates our hypothesis that applying complex neural networks to healthy, static data is mathematically inefficient, reinforcing the necessity of our gating mechanism. Conversely, in the lower RUL regions where active degradation is present, the Probabilistic LSTM demonstrates exceptional tracking of the true RUL with tightening uncertainty bounds.

\subsection{Overall Probability-Weighted Predictions}

\begin{figure}[t]
  \centering
  \includegraphics[scale=.3]{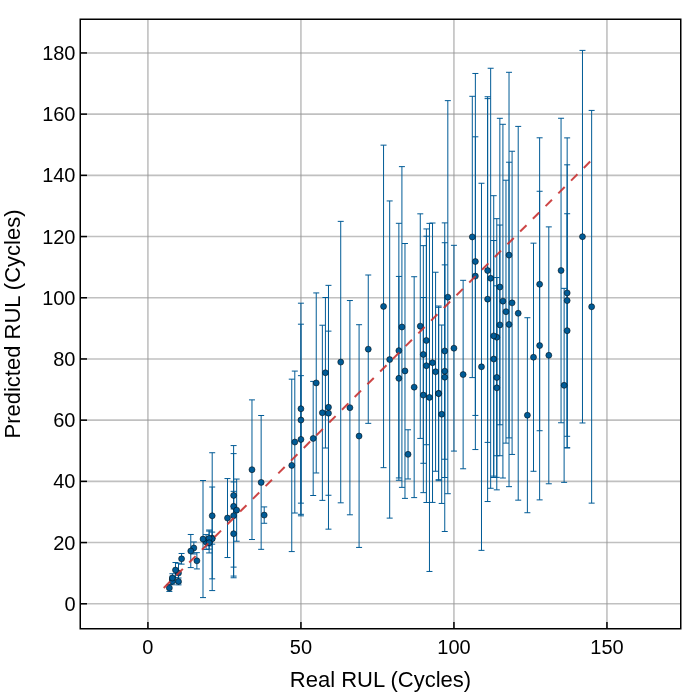}
  \caption{Final weighted RUL predictions on the FD001 dataset. Error bars represent the calculated standard deviation $\sigma$.}
  \label{fig:predictions}
\end{figure}

This section presents the unified RUL predictions, formulated as a probability-weighted ensemble. For any given testing window, the final RUL and its associated variance are dynamically calculated using the continuous degradation probability provided by the autoencoder, ensuring a smooth transition between the statistical Weibull baseline and the deep learning predictions. Figure \ref{fig:predictions}  showcases these finalized results for the FD001 dataset.

The ensemble results demonstrate that the neural network's inherent lack of predictive resolution in the early healthy zone is effectively compensated by the statistical rigor of the Conditional Weibull Survival Analysis. More importantly, the generated uncertainty bands accurately reflect the physical realities of equipment degradation. During the early stages of a trajectory, where no observable degradation indicators exist, the uncertainty is rightfully large, reflecting the population variance at this stage. As the engine deteriorates, the autoencoder shifts the prediction weight heavily toward the Probabilistic LSTM. Consequently, the uncertainty bounds visibly narrow, providing precise predictions with strong confidence at the critical end-of-life stages.

\begin{table}[t] \small
    \begin{center} 
    \caption{Comparative performance across all C-MAPSS datasets (FD001-FD004).}
        \label{tab:metrics}
        \resizebox{1.0\linewidth}{!}{%
        \begin{tabular}{l|l||c|c|c|c}
        \hline
        \textbf{Metric} & \textbf{Model} & \textbf{FD001} & \textbf{FD002} & \textbf{FD003} & \textbf{FD004} \\ \hline \hline
        
        \multirow{3}{*}{RMSE} 
         & Weibull & 37.8 & 40.1 & 60.4 & 60.0 \\
         & LSTM   & 30.8 & 38.4 & 31.2 & \textbf{33.8} \\
         & Ensemble & \textbf{20.9} & \textbf{30.5} & \textbf{24.3} & 46.4 \\ \hline
        
        \multirow{3}{*}{MAE} 
         & Weibull & 32.4 & 33.6 & 53.4 & 50.4 \\
         & LSTM   & 22.1 & 25.5 & 22.6 & \textbf{24.6} \\
         & Ensemble & \textbf{15.0} & \textbf{22.4} & \textbf{15.7} & 35.1 \\ \hline
        
        \multirow{3}{*}{Score} 
         & Weibull & 10,521 & 50,674 & 649,403 & 1,838,029 \\
         & LSTM   & 2,726 & 109,970 & \textbf{3,196} & \textbf{17,383} \\
         & Ensemble & \textbf{774} & \textbf{24,671} & 38,431 & 482,575 \\ \hline
        
        \multirow{3}{*}{PICP} 
         & Weibull & 0.67 & 0.66 & 0.73 & 0.65 \\
         & LSTM   & 0.24 & 0.81 & 0.65 & \textbf{0.79} \\
         & Ensemble & \textbf{0.85} & \textbf{0.81} & \textbf{0.97} & 0.78 \\ \hline

        \multirow{3}{*}{NMPIW} 
         & Weibull & 0.62 & \textbf{0.45} & 1.03 & 0.63 \\
         & LSTM   & \textbf{0.07} & 0.89 & \textbf{0.42} & 0.55 \\
         & Ensemble & 0.46 & 0.46 & 0.77 & \textbf{0.53} \\ \hline

        \multirow{3}{*}{$N/P$} 
         & Weibull & 0.92 & 0.68 & 1.41 & 0.97 \\
         & LSTM   & \textbf{0.29} & 1.10 & \textbf{0.65} & 0.70 \\
         & Ensemble & 0.54 & \textbf{0.57} & 0.79 & \textbf{0.67} \\ \hline
         
        \end{tabular}}
    \end{center}
\end{table}

Results of isolated and ensemble approaches on each dataset are shown in Table \ref{tab:metrics}. Overall results indicate that the performance of the proposed approach declines as dataset complexity increases. Specifically, the Weibull approach struggles when multiple failure modes are present, likely because a single Weibull model is insufficient to fit multiple degradation processes. Concerning the LSTM model, performance decreases when multiple operational conditions are introduced. Nevertheless, integrating the LSTM and Weibull models within the proposed framework generally yields superior results compared to the performance of either model individually, excepting on the FD004 dataset.

Regarding the uncertainty metrics, it is important to note that the Proportional Index metric in isolation can be misleading. For instance, on the FD001 dataset, the Probabilistic LSTM achieves a mathematically lower Proportional Index (0.29) compared to the ensemble (0.54). The LSTM's low ratio is entirely driven by artificially narrow prediction intervals (NMPIW of 0.07), which results in an unacceptably low coverage probability (PICP of 0.24). In practical terms, the true RUL rarely falls within the LSTM's isolated confidence bounds. By contrast, the weighted ensemble dynamically expands the intervals where necessary to achieve a vastly superior and realistic coverage probability of 0.85. Taking all uncertainty metrics into account, the proposed hybrid approach improves uncertainty quantification across all datasets compared to the standalone LSTM and Weibull survival models.

The overall point-estimate performance remains below that of deterministic baselines (RMSE $\sim 8$, C-MAPSS Score $\sim 100$) \cite{asif2022deep}. However, the proposed methodology succeeds in providing uncertainty bounds that faithfully capture the underlying variability of the system. Furthermore, the obtained Proportional Index of 0.54 approaches the performance reported by current state-of-the-art probabilistic models ($N/P \sim 0.45$) \cite{ZHAN2024110383}.

 
\section{Conclusion}

In this paper, we proposed a novel hybrid framework for Remaining Useful Life (RUL) prediction that integrates Uncertainty Quantification (UQ) directly into the prognostic pipeline. By decoupling the prediction task into two distinct operational phases, we overcome the limitations of standard static models. An unsupervised LSTM autoencoder successfully acts as a dynamic health-state classifier, gating the data to eliminate the noise of static, healthy sensor readings. For the early-life phase, a Conditional Weibull Survival Analysis provides a statistically sound, decreasing baseline of expected life. For the active degradation phase, a Probabilistic LSTM trained with a custom Asymmetric Negative Log-Likelihood loss function, employs Monte Carlo Dropout to effectively capture both aleatoric and epistemic uncertainties.

This framework successfully demonstrates that combining deep learning with statistical survival analysis yields highly realistic and safety-critical uncertainty bands. When an engine is healthy and there are no degradation signs, the framework honestly reports high uncertainty, representing true operational unknowns. As degradation becomes active, the uncertainty bounds tighten precisely when maintenance decisions are most critical.

While the calculated metrics remain higher than in state-of-the art approaches, it is important to clarify that the primary objective of this study is not to establish a new state-of-the-art in terms of absolute predictive error ($RMSE$), but rather to introduce a methodological framework that addresses a critical limitation in PHM: the 'flat prediction' bias observed in data-driven models during the early stages of a component's life. By incorporating Weibull survival analysis on the early steps, the proposed hybrid approach improves upon the baseline model's predictions while providing realistic uncertainty intervals that accurately reflect the lack of degradation signals at the start of the trajectory.

Since the proposed hybrid framework constitutes a novel methodological foundation that is still in its early stages of development, it provides substantial opportunities for future refinement through the integration of more sophisticated architectures. Future work will focus on increasing the complexity and capability of the individual modules within the proposed methodology in order to reduce the current performance gap. As a first step, we plan to conduct a rigorous hyperparameter optimization process, including the fine-tuning of the autoencoder probability thresholds and the asymmetric loss penalty factor. In addition, the baseline survival analysis will be extended through the implementation of a dual-Weibull distribution model to independently characterize the time-to-degradation and time-to-failure phases, thereby improving early-stage prognostic predictions. Finally, incorporating mechanisms for degradation-pattern classification could facilitate the extension of the methodology to scenarios involving multiple operational conditions and multiple failure modes.

\section*{Acknowledgment}

This research was partially funded by the Department of Industry of the Basque Government within the Hazitek program ZE-2025/00021.

\bibliographystyle{unsrtnat}
\bibliography{references}  






\end{document}